\pdfoutput = 1
\documentclass[twocolumn]{article}
\usepackage{style}

\usepackage{multirow}
\usepackage{graphicx}
\usepackage{graphics}
\usepackage{amsfonts}

\usepackage{amssymb }
\usepackage{amsmath}

\usepackage[hypcap=false]{caption}

\usepackage{enumitem}

\usepackage{longtable}

\makeatletter
\@namedef{ver@everyshi.sty}{}
\newcommand{\removelatexerror}{\let\@latex@error\@gobble}

\title{Seeing What Shouldn't Be There: Counterfactual GANs for Medical Image Attribution}

\renewcommand\footnotemark{}

\author{Shakeeb~Murtaza\\COMSATS University Islamabad\\}

\newcommand{\ignore}[1]{}

\usepackage{float}

\begin{document}
\maketitle\thispagestyle{fancy}

\maketitle

\rhead{\color{gray}  [S. Murtaza]}

\begin{abstract}
Ascription of an image gives insights into the objects that influence the classification of the whole image or its pixels towards a specific category. These insights help radiologists to visualize deformities in medical imaging. Most of the existing visualization techniques are based on discriminative models and highlight regions of the input image participating in the decision-making of a classifier. However, these approaches do not take all noticeable objects into account as their objective is to classify the input by using a minimal set of discriminative features. To overcome the issue, a counterfactual explanation (CX) based class-oriented feature attribution method is proposed. A counterfactual explanation (CX) explicates a causal reasoning process of the form: ``if X had not happened, then Y would not have happened''. The method is built on generative adversarial networks (GANs) with a cyclical-consistent loss function. We evaluate our method on three datasets: synthetic, tuberculosis and BraTS. All experiments confirm the efficacy of the proposed method. This study also highlighted the limitations of existing counterfactual explanation techniques in producing plausible counterfactual instances (CIs). Accompanying CXs with believable CIs thus provides self-explanatory analogy-based explanations. To this end, a CI generation method is proposed. Also, a novel technique is used to evaluate the quality of CI. The baseline results are produced on the BraTS dataset.
\end{abstract}

\section{Introduction}
\label{sec:1}
The world is already in the industrial 4.0 era, in which artificial intelligence (AI) has an imprint on every prominent field. With the advent of deep learning, AI has surpassed human intelligence in certain fields, especially in computer vision (CV). A main hurdle in the adoption of deep learning is its black box behaviour. Certain techniques have been devised already to make these models interpretable. In this section, background on deep learning and interpretability methods is presented. 

\subsection{Deep Learning}
Deep learning is a branch of machine learning (ML) which provides state-of-the-art solutions in different areas of CV. These advances are greatly impacting medical image diagnostics (MID) with the development of tools and systems. Despite enormous performance-based breakthroughs, the lack of interpretability hinders the reliance on these techniques in MID-type safety-critical domains.  

Deep learning deals with learning hierarchical representations or features of raw data. Typically, deep learning is performed with artificial neural networks (ANNs). ANNs are based on a parallel processing model that is inspired by the biological neuron, which is the core component of the human brain. The brain is an assemblage of gigantic neurons that collaborate to process information from different senses and define actions in response to that information. Just like a brain’s neuron, ANNs consist of different perceptrons (basic building blocks of neural networks). A simplest neuron network consists of an activation function, a set of weights, and bias represented as $\sigma$, w and b, respectively. Single perceptron with some inputs and respective weights is shown in Figure \ref{fig:sing_multi_percep} (a). Mathematical representation of a perceptron with ``n” inputs is given below.
\begin{equation}
	\sigma\left(\sum_{i}^{n}\left(w_i\right)\ +\ b\right)
\end{equation}
The perceptrons are stacked together to build layers, and these layers are then stacked to build a deep neural network, as shown in Figure \ref{fig:sing_multi_percep} (b). At the output of each layer of perceptron, the activation function scales the output value in a particular range as per its characteristic \cite{intro_5}.
\begin{figure}[!h]
	\centering
	\includegraphics[clip,width=1\linewidth]{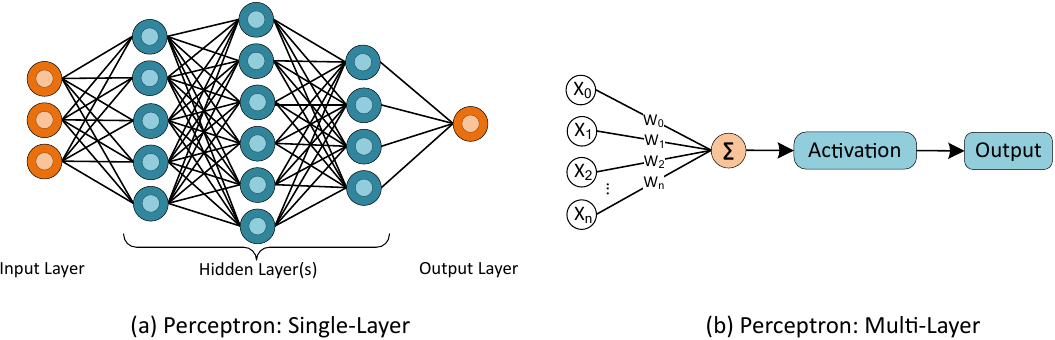}
	\captionsetup{justification=centering}
	\caption{Single and Multi Layer Perceptron}
	\label{fig:sing_multi_percep}
\end{figure}
Studies have established that deep ANN learns hierarchical features that are more effective than low-level features of the swallowed neural network \cite{intro_22}. However, these networks are very hard to train due to the large number of connections in the network, and these are not feasible in high-dimensional tasks such as computer vision. To cope with this issue, convolutional neural networks (CNNs) are introduced. Instead of a dense connection, CNN learns a special relationship by convolving a kernel on the input image, as shown in Figure \ref{fig:cnn}. At each layer, a set of filters is convolved, and an activation function is applied to generate output. CNNs have turned out to be state-of-the-art models in many computer vision tasks, e.g. image classification, segmentation, localization, object recognition and detection \cite{intro_22, intro_6, intro_23, intro_24, intro_25}. Despite many performance breakthroughs, deep learning is notoriously known for its black-box nature, which hinders its applicability in safety-critical applications such as medical MID. The objective of this research is to devise a method for enabling deep learning to be interpretable. The interpretability is introduced in the Section \ref{Sec:interpretability}.
\begin{figure}[!h]
	\centering
	\includegraphics[clip,width=1\linewidth]{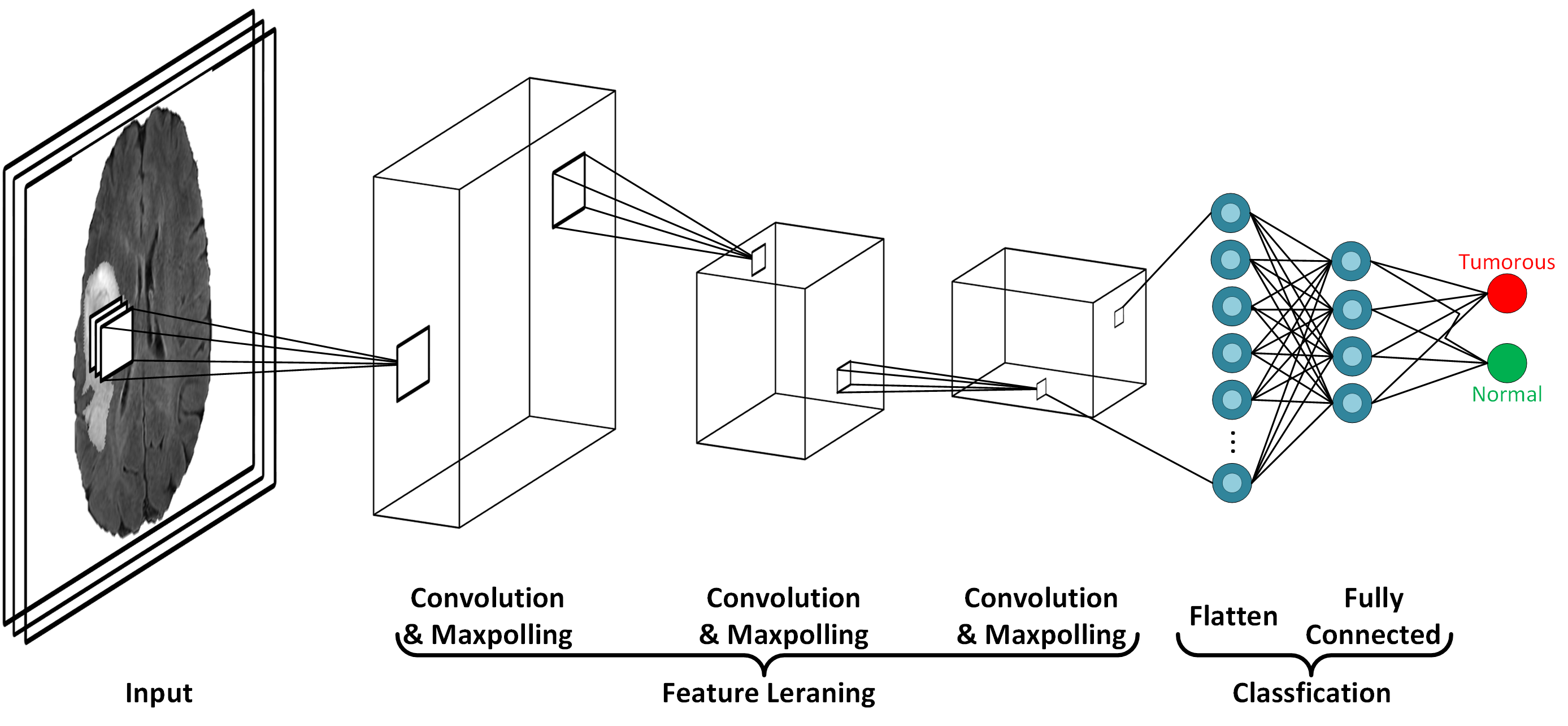}
	\captionsetup{justification=centering}
	\caption{Illustration of Convolution Neural Network}
	\label{fig:cnn}
\end{figure}
\vspace{-0.5cm}
\subsection{Interpretability}\label{Sec:interpretability}
\begin{figure}[!h]
	\centering
	\includegraphics[clip,width=1\linewidth]{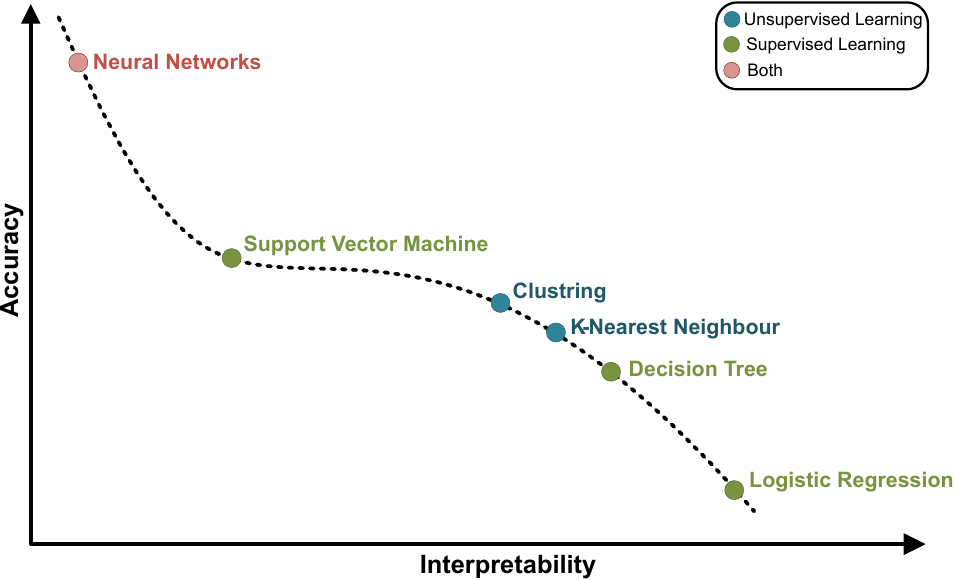}
	\captionsetup{justification=centering}
	\caption{Accuracy vs Interpretability}
	\label{fig:acc_vs_inter}
\end{figure}

The interpretable or explainable model is a computational model whose decisions could be understandable to humans and justified logically \cite{intro_7, intro_8}. Although there is no universally agreed-upon single definition of interpretability, different researchers have defined it differently. One of the best definitions provided by an interdisciplinary conference known as FAT (Fairness, Accountability and Transparency) to ensure that the decisions of the model, along with the data that drives those decisions, are explainable to end-users, even if they are non-technical \cite{intro_12}. In \cite{intro_9}, the researcher states that a model is interpretable if it provides trust, causality, transferability, and informativeness to make fair and ethical decisions. A greater extent of interpretability helps us in understanding the reason for a prediction more easily. Interpretability level is deliberated according to the understanding of the prediction by a human being. In recent years, much work has been done in the context of interpretable ML, particularly to address the interpretability and accuracy trade-off \cite{intro_1, intro_2} as shown in the Figure \ref{fig:acc_vs_inter}.

\subsubsection{Importance}
The outstanding performance of deep learning-based systems does not ensure the reliability of the model. They are outperforming traditional ML algorithms in certain tasks, but they can't be trusted unless they provide some human-understandable explanations. Without understanding the decision-making process completely, trusting these models can cost us a lot \cite{intro_10}. Providing justifications or explanations of a model’s decision is vital to gain end-user trust.

The dire need for interpretable models has been raised in recent years, particularly when a model which performed very well on training data performs poorly when deployed in the real world. For example, in \cite{intro_10}, a military classifier was used to distinguish between friendly and enemy tanks. It produced excellent results on training and validation data, but performed poorly in a real scenario. Upon investigation, it was found that pictures of friendly tanks were taken on a sunny day, whereas that of enemy tanks was taken on a cloudy day. So, instead of learning the desired features (differentiating between the two tank types), it was classifying on the basis of more prominent features (environmental conditions).  A similar situation was faced by the researcher in \cite{intro_11}, when they used a classifier to separate wolves from husky dogs. The classifier was classifying wolves on the basis of background snow rather than other features. In 2016, a study of correctional offender management profiling for alternative sanctions (COMPAS), a widely used criminal profiling tool, showed that the predictions were biased and unreliable \cite{intro_12}. Besides these issues, the European parliament adopted the General Data Protection Regulation (GDPR) \cite{intro_13} in 2018, according to which a meaningful explanation is needed, especially when the decision is made using automated systems, due to which it became essential for such black-box systems to explain the decisions to make the process transparent \cite{intro_14, intro_15}. 

One problem with conventional deep learning methods is that they are normally evaluated using much simpler matrices, e.g. classification accuracy, which provide partial information about efficient use of these models for real-world tasks. Besides these evaluation criteria, the model must be capable of learning and differentiating between different objects like human beings.

In contrast to low-risk environment applications (e.g. movie recommender), critical applications are required to produce the reasons for the prediction as well as the predicted output.  In such scenarios, the model must interpret why the model comes to this decision \cite{intro_7}.

\textbf{Human learning: }We have a model of this world in our brain. When an accidental situation occurs, our mental model tries to justify the occurrence of the event by giving an explanation for the unexpected event. For example, I felt pain every time in my throat when I inhale the fragrance of different products. The same is the case with deep learning when used in scientific applications; scientific exploration will be stuck if the models are unable to explain the reasons for their predictions. Normally, humans do not care about explanations until an unexpected event occurs. For example, why is my model taking more memory than usual while training?

Researchers and scientists from different domains start working on deep learning to automate and speed up research tasks. The primary intention behind scientific research is to gain knowledge and use it for the benefit of humankind. Many of the problems are solved by utilizing deep learning (black box) models and large datasets. These models encompass the knowledge base rather than data and interpretability, which helps us in extracting the knowledge base of the model.

Deep learning models may learn biases from training data, and this makes the model racist against specific categories. Interpretability is a valuable tool for debugging and detecting bias in the output. The deep learning model that is trained for detecting cars vs no-car in the image may point out a common feature in all positive images rather than cars. For example, if all positive class instances contain a logo of a car, then the model is more likely to learn the logo in the image, as this is the most common feature in all images with a car.

These models can be debugged and audited only if they are interpretable. Interpretability is a technique not only helpful in the research and development phase, but it is also helpful after deployment. If anything goes wrong after deployment, then an interpretable model helps us to understand the reason for the wrong output. These interpretations produce valuable insights to fix the system. In \cite{intro_11}, the author discussed an example of a wolf versus a husky classifier that misclassifies some huskies as wolves. By utilizing interpretable models, it is inferred that this misclassification occurred due to snow on the image.

Interpretability models are broadly classified into two categories, ``ante-hoc'' and ``post-hoc'' \cite{intro_16}. Ante-hoc models are inherently interpretable due to their simple structure and glass-box approaches; examples include linear regression, decision trees, and fuzzy inference systems. However, post-hoc models are not inherently interpretable. Other models or methods are used to explain the decision of the system, rather than explaining the model itself, i.e., Local interpretable model-agnostic explanation (LIME) \cite{intro_11}. Another point that distinguishes post-hoc methods from prior methods is that they are applied after training the model.

\subsubsection{Types of Ante-hoc interpretability}
In this section, different types of ante-hoc interpretability \cite{intro_17} are introduced.

\textbf{Text explanations:}
This type of interpretable model deals with providing textual explanations of model decisions. Typically, the explanations can be generated by training two networks simultaneously. One is trained for prediction, and another network (e.g., a recurrent neural network) is trained to generate the explanations. An example of a text-based method is to train an image-to-caption network using representation learning along with the classification network that explains features of the image \cite{intro_26, intro_27}.

\textbf{Visualizations:}
This type of model provides interpretation by synthesizing a visualization to determine the features that are learned by the model. An example of this technique is t-SNE that reduces the input features and visualizes them on the 2d plane. In \cite{intro_28}, the author explains the reason for the output by tweaking the input until a specific prediction was achieved. In an example, the author considers an image full of noise and changes it until the network labels this image as bananas. Different techniques for visualizing the feature map of CNN are presented in \cite{intro_29} that allow us to visualize the activation map at any hidden layer.

\textbf{Local Explanations:}
Explanation of the whole mapping learned by the network is very difficult to obtain. To cope with this issue, some authors proposed local explanation methods. These methods calculate the gradient of the input with respect to the predicted class. The output of such methods is then overlaid on the input image for computing saliency or a heat map. However, the output of such methods may be misleading due to an unexpected feature in a specific instance. Examples of the methods include CAM \cite{intro_30}, grad-CAM \cite{intro_31}, and methods that leverage pseudo-label sampling for finer localization \cite{murtaza2023discriminative, murtaza2023dips, murtaza2024leveraging, murtaza2025ted}, alongside the realistic evaluation protocols for these methods \cite{murtaza2025realistic}.

\textbf{Example-Based Explanations:}
Example-based explanations are defined as a function that assigns weights to each data point of input based on their influence in a particular decision \cite{intro_32}. These types of explanations are produced individually for each input and provide useful insights that help in understanding the behaviour of the model. Example-based explanations are only feasible if the instances of the data can be represented in an understandable way. In contrast to tabular data with thousands of features, images can be represented directly in an understandable way for human beings. These explanations help us to understand the knowledge in a contrastive way that is inferred by the model from training data. Let’s understand the concept of example-based explanations from examples \cite{intro_7}.

\begin{itemize}
	\item When a physician observes a patient with fever or cough, these symptoms remind him of the condition of his past patient with similar symptoms. From these analytics, physicians suggest particular tests/medications to the patient.
	\item Let’s say a data scientist is working on a client’s project for risk factors analysis, which causes the failure of the keyboard production machine. This data scientist reuses code extracted from his previous projects because he infers from requirements that his client expects the same analysis that he has done on other projects.
\end{itemize}

These examples elaborate the human thinking strategies through examples in the form; If Feature A is similar to feature B that caused X, then humans can predict that A causes X. This way of contrastive thinking is also known as counterfactual thinking. In explainable AI, counterfactual explanations (CX) are just like counterfactual thinking and fall under the category of example-based explanation. CX is described briefly in the next sections.

\subsubsection{Counterfactual Explanation (CX)}
\begin{figure}[!h]
	\centering
	\includegraphics[clip,width=0.8\linewidth]{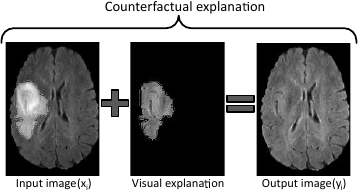}
	\captionsetup{justification=centering}
	\caption{Counterfactual Explanation}
	\label{fig:coutnerfactual_exp}
\end{figure}

CXs are contrastive and useful to describe the reason for a particular output instead of global explanations \cite{intro_7}. The process of CX can be defined in the form: If A was A*, then the outcome should be X rather than Y \cite{intro_33}. In explainable AI, CXs describe changes in the current input that make this input an instance of another class. An illustration of CX is shown in Figure \ref{fig:coutnerfactual_exp}.

Humans tend to think in a counterfactual way. For example, if a person's loan application was rejected, he would be interested in finding the accepted version of his application vs the rejected version \cite{intro_7}. In a similar way, a physician can ask, ``Why didn't the drug work for a patient, and he is interested in finding a patient with similar conditions, on whom the particular drug worked''. In this way, humans try to understand the underlying mechanism of the system to make a certain decision. From these examples, it is shown that CXs are easier for humans \cite{intro_7}. So, the focus of this work is on counterfactual-based explainability.

Existing state-of-the-art methods for CXs only describe the smallest region that leads the classifier to change its results. These techniques cover the feature with some pre-defined values, and the resultant counterfactual is implausible. To cope with these issues, a generative adversarial network (GANs) based method is proposed that could generate plausible counterfactual instances (CIs) and measure the change that leads towards the CX. These CXs help us to explain the reason for the decision and identify the abnormalities in the image.

\subsection{Motivation}
\begin{figure}[!h]
	\centering
	\includegraphics[clip,width=7cm]{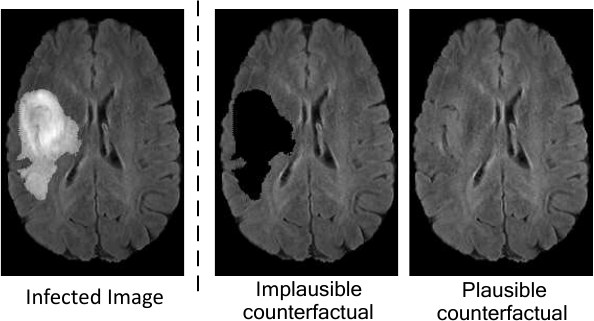}
	\captionsetup{justification=centering}
	\caption{Plausible vs Unplausible Counterfactual Explanation}
	\label{fig:paus_unplaus}
\end{figure}

A CX explicates a causal reasoning process of the form: ``if X had not happened, Y would not have happened'' \cite{intro_7}, for example, ``if I hadn’t had these symptoms, I would not have this disease''. Existing de facto state-of-the-art CX methods tend to describe the smallest changes to the input features that alter the prediction of a classifier \cite{intro_34, intro_35}. Although such methods delineate vital discriminatory features, the resultant CI is not semantically plausible. An example of a plausible and implausible counterfactual instance is depicted in Figure \ref{fig:paus_unplaus}. Accompanying CXs with believable CIs thus provides self-explanatory analogy-based explanations. For example, in medical diagnostics, it is useful to address the question ``why is this particular disease diagnosed?'' by providing an analogy between the input (i.e. ``how a scan looks'') and a relevant CI (i.e. ``what it should look like'').

Previous CI-based CX techniques replace a part of an input image (e.g. a square tile) with a specific region of a counterfactual image (i.e. a CI) \cite{intro_33}. Such techniques thus intervene in the original data space, but make only a restricted number of changes. Consequently, the generated CXs are not plausible. Furthermore, these techniques require a dataset of pairs of images from the input and counterfactual classes, which may not be readily available. Recently, generative models have been used to learn transformations that, when applied to an input image, produce counterfactual images \cite{intro_36}. Since this is a one-to-many mapping, specific counterfactual images must be generated randomly from the input image over the transformation set. However, this under-constrained mapping can lead to the production of irrelevant CXs, reflecting undesirable discrepancies between the images.

\subsection{Contributions}
This paper addresses the interpretability/explainability of deep learning, particularly in the MID domain. Overall, this work focuses on devising a CX method for MIDs. In contrast to existing CX methods, the main objective of this work is to explain model decisions using a plausible counterfactual. The main idea is to generate counterfactuals of a given input image and then measure the difference between the original image and the generated counterfactual. Highlights of main contributions are as follows:

\begin{itemize}
	\item Pose a novel problem of plausible counterfactual generation for CX.
	\item Develop a method for CX against plausible counterfactuals. 
	\item The method is tested on a synthetic and two medical imaging datasets, including tuberculosis and the BraTS dataset, and compared with existing methods.  
	\item Propose a method to evaluate the quality of generated counterfactuals and provide baseline results. 
\end{itemize}

\subsection{Organization}
The rest of the paper is organized as follows: the problem statement and literature review are described in section \ref{sec:2}. Section \ref{sec:3} covers the problem statement along with the literature review. The proposed methodology, including the results, is described in section \ref{sec:4}. At the end, the conclusion and future work are presented in sections \ref{conclusion} and \ref{futurework}, respectively.

\section{Literature Review and Problem Statement}
\label{sec:2}
In this section, a literature review is presented, which is required to understand the context of this research. Literature review for interpretability-based methods and image-to-image-translation methods is presented in Section \ref{literature}. The problem statement is presented in Section \ref{problem_statement}.
\subsubsection{Literature Review}\label{literature}
Neural networks have outperformed human beings in certain tasks such as object recognition \cite{intro_24}, object tracking \cite{rw_0}, etc. Deep learning becomes an essential tool for vision-based applications due to its tremendous predictive accuracy. However, in critical applications, it is essential to exploit the reason for an output along with the results. Methods for interpreting the model become an essential part of a robust model validation process. Interpretability is very important in critical applications (e.g. medical) where the learning of appropriate features must be guaranteed before the actual implementation of the model so that people can trust the model. 

Techniques for model interpretability are being adopted for knowledge exploration and analysis. The model learns features from training data and becomes the source of knowledge. So, the insights of a model must be interpretable to use the inferred knowledge for scientific discoveries.

This section mainly focuses on the most commonly used interpretability techniques in DL (e.g. CAM, grad-CAM, etc.) and the techniques for image-to-image translation.

\textbf{Interpretability Methods.} The techniques for interpretability are broadly classified into two categories; discriminative models-based methods and generative models. Different methods are built upon discriminative models to highlight class-specific features \cite{intro_30, intro_31, rw_2, rw_3, rw_4}. The most commonly used technique for visualizing class-oriented features is class activation mapping (CAM).

\begin{figure}[!h]
	\centering
	\includegraphics[clip,width=1\linewidth]{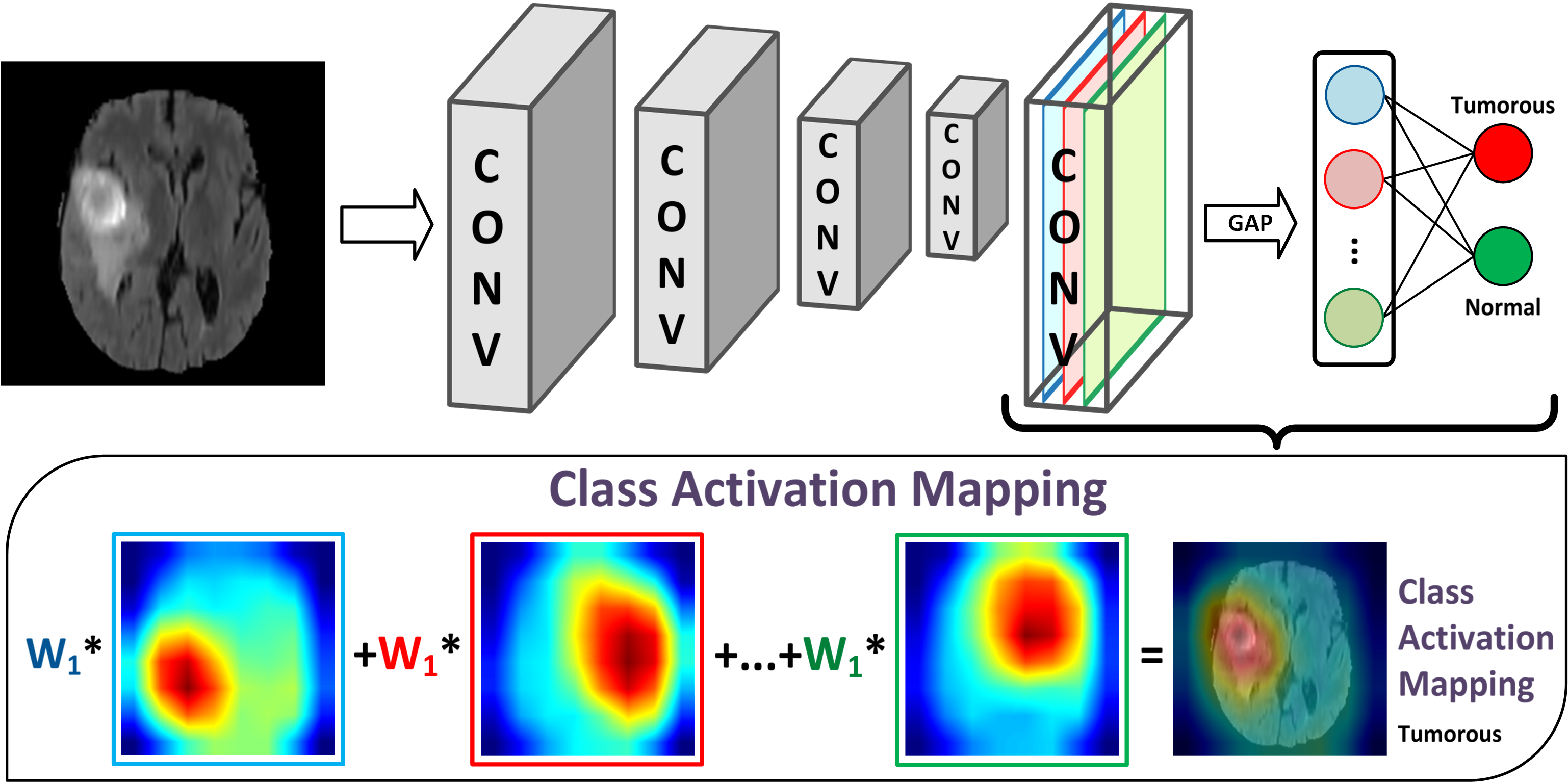}
	\caption{Illustration of CAM method}
	\label{fig:camIllustration}
\end{figure}

In CAM, the flattened layer is replaced with a global average pooling (GAP) layer after the last convolution layer in the network and is connected to a dense output layer. Class-specific weights of the output layer are then multiplied with the output of the last convolution to produce a heat mask as shown in Figure \ref{fig:camIllustration}. More specifically, output of last convolution for kernel ``k" at spatial location $(x, y)$ is represented by $f_k(x, y)$ and the result of GAP $(F^k)$ is calculated as $\frac{1}{n}\sum_{x, y}{f_k(x, y)}$. The class activation maps are produced by utilizing the output of the last convolution layer and class-specific weights $(w_c^k)$ of the output layer using $\sum_{k}{w_c^kf_k(x,y)}$. The generated feature is resized and overlaid on the original image to produce a heat map \cite{intro_30}.

However, CAM helps to visualize features learned by the last convolution layer. In order to visualize the feature maps at every layer, Grad-CAM \cite{intro_31} was introduced. To achieve this, backpropagation is employed at the targeted layer in the network with respect to the output class as shown in Figure \ref{fig:camIllustration}. For this purpose, we must compute the gradient of class C with respect to the activation maps $A^m$ at a selected hidden layer using this equation \cite{intro_31}:
\begin{equation}
a_m^c\ =\ \frac{1}{z}\sum_{x}\sum_{y}\frac{\partial y^c}{\partial A_{x,y}^m}
\end{equation}

These values are multiplied by the feature maps, followed by a rectified linear unit (ReLU) to obtain the activation map as given in equation \ref{eq:2.2}.
\begin{equation}\label{eq:2.2}
\mathcal{L}_{grad-CAM}=ReLU\left(\sum_{m}{a_m^cA^k}\right)
\end{equation}
The size of the generated feature map is dependent on the output size of the selected convolution layer. So, this map must be resized to overlay on the original image. 

\begin{figure}[!h]
	\centering
	\includegraphics[clip,width=1\linewidth]{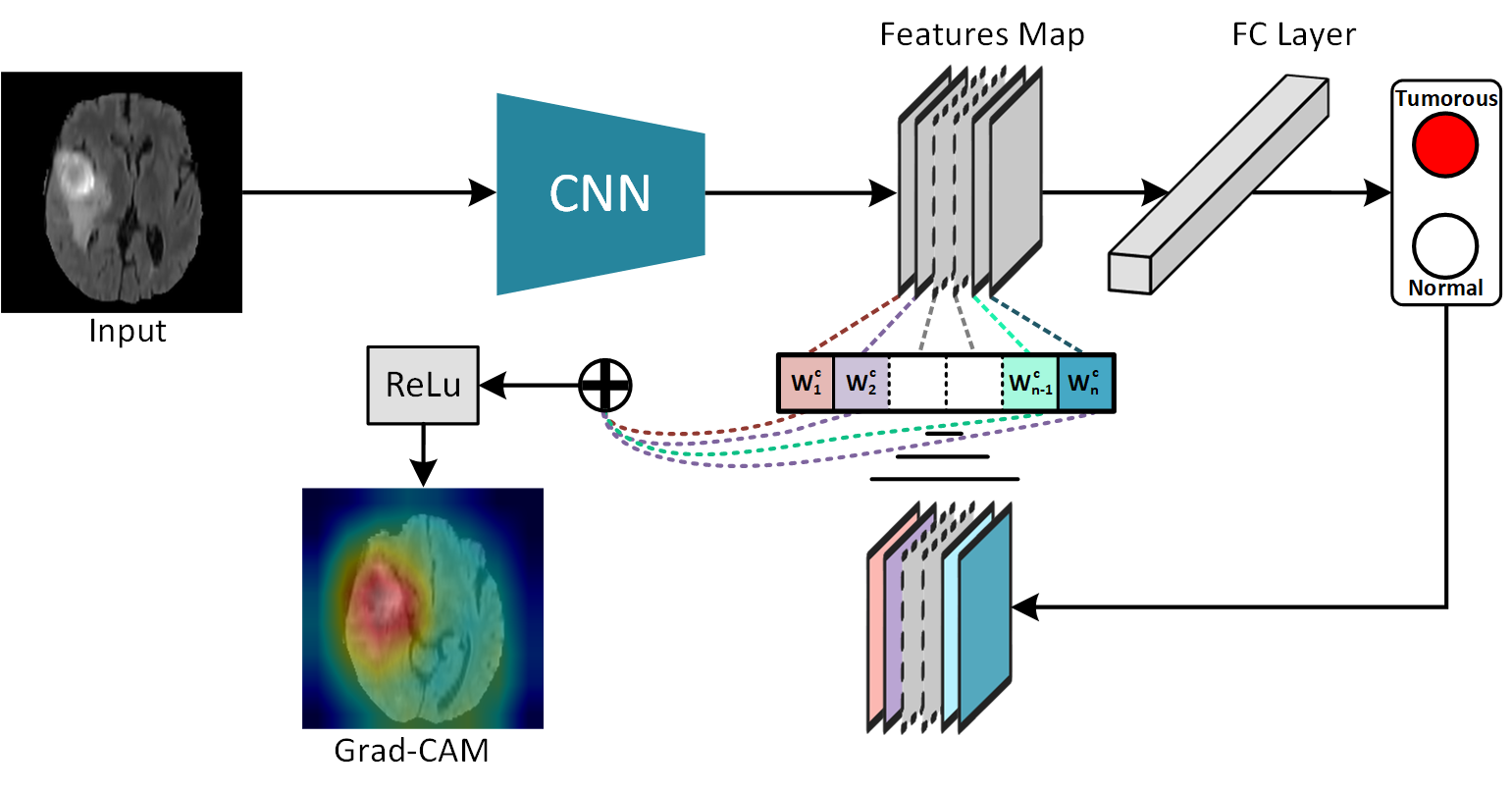}
	\captionsetup{justification=centering}
	\caption{Illustration of Grad-CAM}
	\label{fig:grad_camIllustratio}
\end{figure}

\begin{figure}[!b]
	\centering
	\includegraphics[clip,width=1\linewidth]{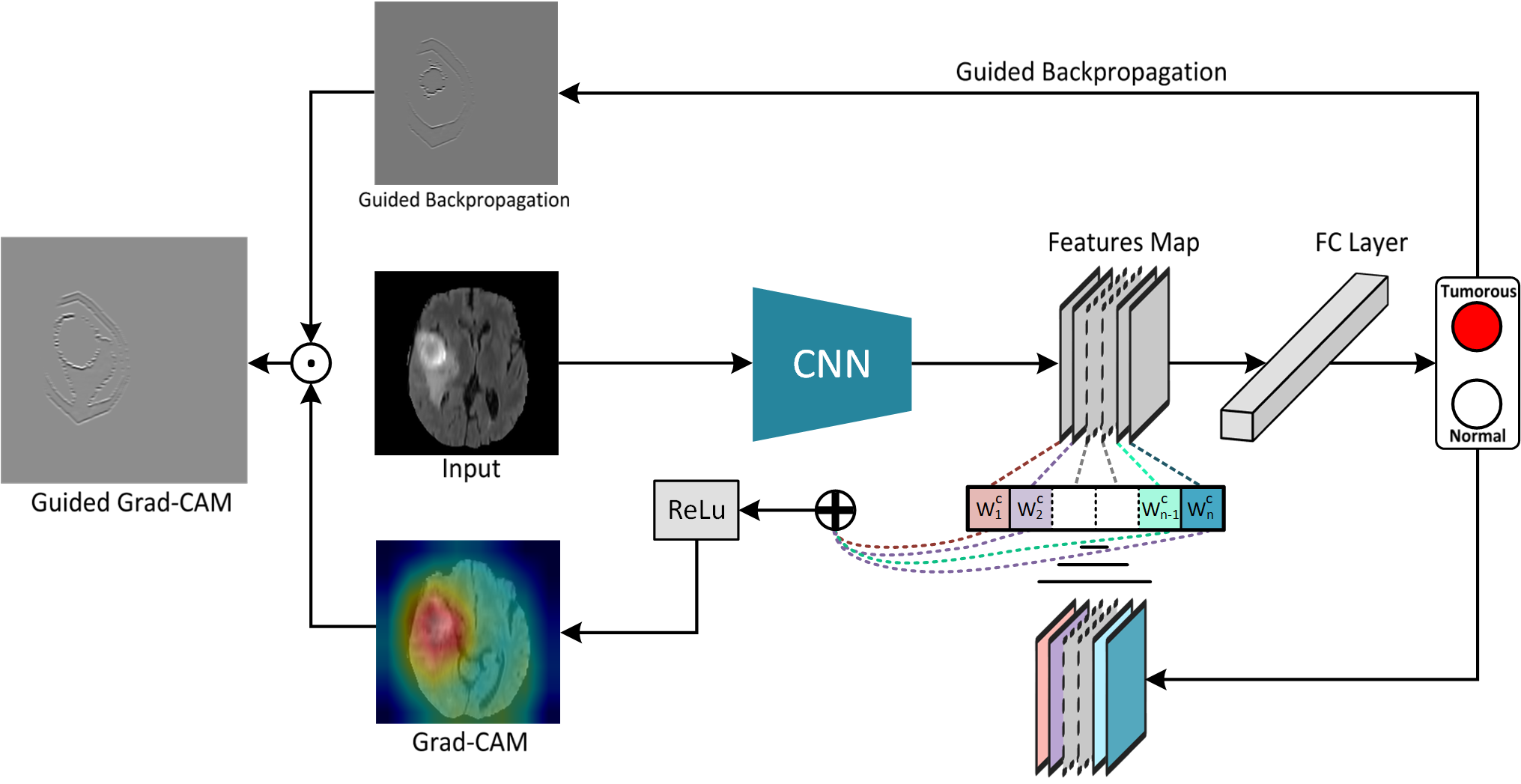}
	\captionsetup{justification=centering}
	\caption{Illustration of Guided Grad-CAM}
	\label{fig:guided_grad_camIllustratio}
\end{figure}

To produce the final feature map, Grad-CAM resizes the generated feature map and produces a low-resolution map. To cope with this issue, guided backpropagation was employed \cite{intro_31}. Map generated by guided-back-propagation and grad-CAM is combined through point-wise operation for producing smooth feature map as shown in Figure \ref{fig:guided_grad_camIllustratio}.

The CAM technique has been recently used in many studies on medical images, including the detection of pulmonary nodules from CT scans \cite{rw_7}, recognition of skin diseases \cite{rw_8}, localization of diabetic retinopathy lesions \cite{rw_9}, and visualization of tuberculosis \cite{rw_6}. An issue with employing an activation map for explaining class-specific features is that the output of each value is determined independently, and the generated map is not smooth. To resolve this issue, the activation atlas technique is proposed to visualize average activation values rather than single input values \cite{intro_29}. In \cite{rw_6}, deconvolutional layers are stacked on top of the convolutional layers to expand discriminative features across an image for a smooth feature map of tuberculosis from chest X-rays.

Another class of visualization techniques deals with creating saliency maps by backpropagating the gradients back to the input image. It involves different methods, including excitation back-propagation \cite{rw_17}, guided back-propagation \cite{rw_12} and integrated gradients \cite{intro_20}. Related techniques have been utilized in medical image domain, including the proposal of saliency technique for pinpoint lumbar degradations \cite{rw_13}, localization of fetal anatomy based on guided back-propagation like technique \cite{rw_14, rw_15}. A similar technique is also used for localizing the fetal heart \cite{rw_16}.

Other techniques for explaining the output of the model use CXs, many of them use irregular and meaningless values to be replaced with different portions of the image to alter the model’s prediction \cite{intro_20, intro_18, intro_19, intro_21}. The process of replacing image regions iteratively for counterfactual explanation is proposed by \cite{intro_34}. These meaningless values lead the model to generate implausible results. In \cite{intro_35}, a technique to extract the minimal region is devised, based on which the model makes a particular prediction. 

It is worth noticing that the CAM-based approach relies on classification that may not visualize the full details of the object of interest. Furthermore, CAM deals with analyzing the last feature map of the network, which requires post-processing of the network prediction. To cope with the issues, a GAN-based approach is introduced in \cite{intro_36}. The approach creates CX explicitly as a function of the image without relying on classification. This problem is posed as an image translation problem, and the mapping function learns to map an image of a specific class to any other image of the baseline class. As this function is under-constrained, its flexibility to map the same set of images to any random permutation of images leads to the creation of false positives merely due to anatomical discrepancies between the input and translation images. To cope with this issue, a new method is proposed, stated in section 3, to learn CI generation along with the CXs.

\textbf{Image-to-Image translation.} The intention behind image-to-image translation is to learn a mapping between the input and output images. Examples of image-to-image translation include image denoising, resolution enhancement, face swap, etc. For image-to-image translation, numerous feedforward CNNs are proposed. These networks are trained and updated using backpropagation. For semantic segmentation, the network generates labels from the input image. Different networks, such as U-Net, SegNet, FCN, etc., are being used for segmentation.

Certain GAN-based techniques are used for image-to-image translation. GAN consist of two networks; Generator and Discriminator. The generator is responsible for generating an image that belongs to a specific distribution, and the discriminator helps the generator in the learning process by playing a min-max game. Conventional GANs accept random noise and generate images that belong to the distribution of training images \cite{rw_26}.

In conditional GAN (cGAN), an additional supervisory signal (condition ``c'') is provided along with the noise to produce images with certain features. This condition is passed to both generator and discriminator. For example, if we want to generate handbag from its outlined image then we need to provide colour of the handbag as condition \cite{rw_27}. As a cGAN accept random noise with a condition, it does not guarantee the exact same results each time. To tackle this issue, Pix2Pix GAN was introduced for translation between two domains. It requires input x along with some output image y, and it learns a mapping function $f:x\rightarrow y$. However, this approach is very useful for translating images between two different domains but requires input-output pairs \cite{rw_28}. These pairs are not always present, e.g. in the medical domain (normal MRI against anomalous MRI of the same patient).

Cycle-consistent GAN (Cycle-GAN) was introduced to overcome this input-output pair issue. It learns translation between input (x) and output (y) by simultaneously learning two functions $g:x\rightarrow y$ and $f:y\rightarrow x$. In this method, x is translated to y, and then y is translated back to the original domain x, to measure the quality of translation \cite{rw_29}.To take advantage of unpaired translation using cycle-GAN, a method is proposed for CX by generating a change map between the generated CI and the input image.

\subsection{Problem Statement}\label{problem_statement}
Although deep learning has achieved extraordinary competence in many computer vision tasks, but are complex and opaque. To overcome the opacity of deep learning methods, various visual explanations method has been deployed recently \cite{intro_30, intro_31, rw_2, rw_3, rw_4}. Most of the existing methods focus on explaining classifier decisions by highlighting areas of the input image that mostly contribute towards the decision \cite{intro_30, intro_31}. These methods are neither counterfactual nor contrastive. Such casual reasoning beyond correlation is crucial for full interpretative explanation of decisions \cite{intro_34, intro_35}.
Although a few counterfactual techniques have recently been proposed, they require pairs (i.e. input, counterfactual image) to explain input images with respect to counterfactual images. However, it is not always possible to have an image pair in domains such as medical imaging. To overcome the issue, a visual feature attribution GAN (VA-GAN) model was proposed \cite{intro_36}. The VA-GAN produces explanations of the input image with respect to any permutation of the image in the counterfactual domain. However, due to semantic unalignment between the input and the counterfactual image, the model produces false-positive explanations. The aim of this work is to tackle the situation where input images along with their CIs are hard to obtain. 
Additionally, the generated CIs of the existing models are not plausible \cite{intro_34, intro_35, intro_36}. To produce a plausible CI is vital for providing self-explanatory analogy-based explanations. To this end, the objective of this work is to develop a CX method for producing plausible CIs along with CX.

\section{System Architecture and Proposed Methodology}
\label{sec:3}
This section presents proposed counterfactual instances and explanation generation methods. The task of CI generation is posed as unpaired image-to-image translation, and CX as image-to-image conversion mapping. A technical background on GANs and image-to-image translation is provided in Section \ref{arch_background}. Methodological approach, proposed model architecture and model training are described in Section \ref{methodology}.

\subsection{Architectural Background}\label{arch_background}
This section provides the fundamental details of the methods that are utilized for the development of our proposed method.

\subsubsection{Generative Adversarial Networks (GAN)}
\begin{figure}[!h]
	\centering
	\includegraphics[clip,width=1\linewidth]{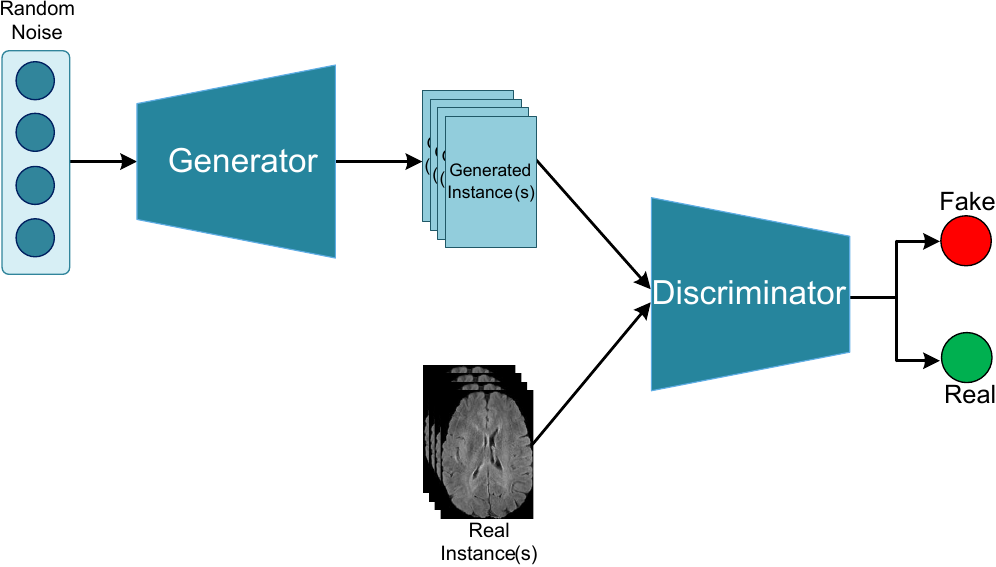}
	\captionsetup{justification=centering}
	\caption{Generative Adversarial Networks}
	\label{Fig:1}
\end{figure}

GANs fall under the category of generative models. These are state-of-the-art networks for generating images whose distribution must be same as the training data. It comprised of two networks; one is the generator ``G", and the other is the discriminator ``D" as depicted in Figure \ref{Fig:1}. Discriminator network helps the generator in the learning process by tagging its output as \textbf{fake} or \textbf{real}. In response to discriminator feedback, the generator tries to produce an image that belongs to the distribution of training data \cite{rw_26}. In other words, both networks play a min-max game; D tries to tag G’s output as fake, and the generator tries to fool D, as discussed. The mathematical representation of GANs’ objective is given in equation \ref{eq:1} \cite{rw_26}.
\begin{multline}\label{eq:1}
	{min}_G{max}_DV(D,\ G)\ =\ \mathbb{E}_{x\sim p_{data(x)}}[log[D(x)]]\\ +\mathbb{E}_{z\sim p_{data(z)}}[log[1-D(G(z))]]
\end{multline}
Where x is an instance that belongs to the training set, and z is a random noise vector drawn from a known distribution. In this objective function, D(X) and D(G(x)) correspond to the discriminator’s output on training data and generated output, respectively.

\subsubsection{Cycle GANs for Image-to-Image translation}
\begin{figure}[!h]
	\centering
	\includegraphics[clip,width=1\linewidth]{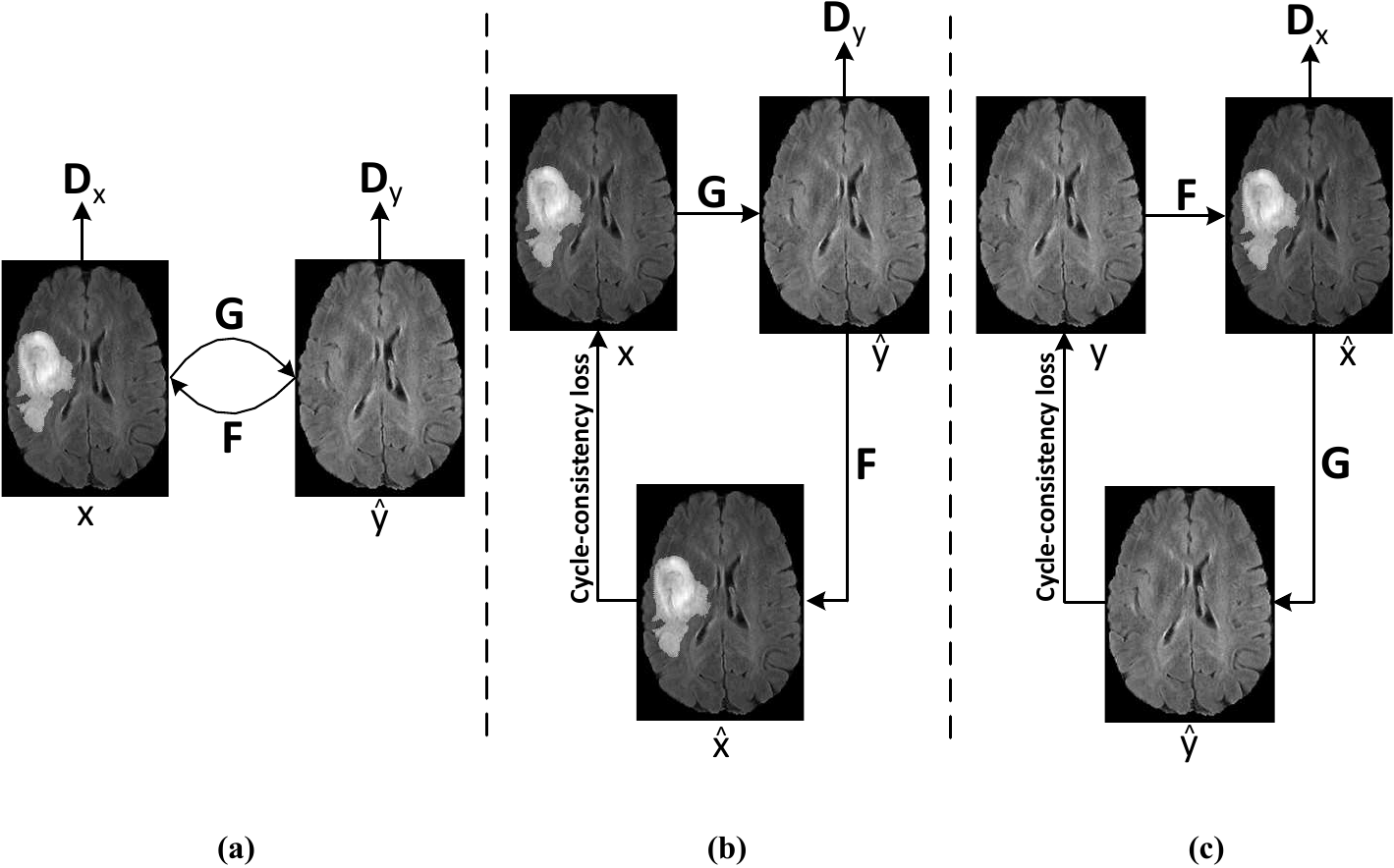}
	\caption{Cycle-GAN: (a) Model consists of two generators ($G:x\to y$ and $F:y\to x$) along with two discriminators, (b) Forward cycle-consistency loss (for reconstruction of input) (c) Backward cycle-consistency loss}
	\label{Fig:2}
\end{figure}

In image-to-image translation, the goal is to translate an image from some source domain to some target domain. Typically, the models are trained in a supervised setting by using a paired dataset. A disadvantage of this approach is that it requires a paired dataset, which may be difficult to acquire/prepare, e.g. in domains such as medical imaging. To overcome the issue, an unsupervised approach, referred to as Cycle-GAN, is proposed to translate images across domains without a paired dataset \cite{rw_29}. Essentially, Cycle-GAN consists of two generators (named as G and F) and two discriminators ($D_y$, $D_x$). First generator G learns translation between input images (x) to the target domain (y) with the assistance of the concerned discriminator. The second generator F learns the inverse of the first generator by translating y to x as shown in Figure \ref{Fig:2}. Following loss is employed for generator ``$G$'' that performs translation from x to y with its discriminator ``$D_y$" \cite{rw_29}.
\begin{multline}\label{eq:2}
\mathcal{L}_{GAN}(G, D_y, X, Y) = \mathbb{E}_{x\sim p_{data(x)}}[1-log (D_y(G(x)))]+\\ \mathbb{E}_{y\sim p_{data(y)}}[log (D_y(y))]
\end{multline}
Here, G learns translation from input images ``x" to target domain ``y", and $D_y$ helps the generator in translation learning by tagging its output. Same loss function is utilized for second generator ``F'' with the assistance of discriminator $D_x$. Loss for generator ``F" is given below.
\begin{multline}\label{eq:3}
\mathcal{L}_{GAN}(F, D_x, X, Y) = \mathbb{E}_{x\sim p_{data(x)}}[1-log (D_x(F(x)))]\\ +\mathbb{E}_{y\sim p_{data(y)}}[log (D_x(x))]
\end{multline}
Although the above given losses ensure the adversarial learning for the network, the learned function must be capable of two-way mapping from x to y and vice versa. In other words, we can say there must be a mechanism to translate the input image to the original form, e.g. $x\rightarrow G(x)\rightarrow F(G(X))\approx x$. This behaviour can be achieved using cycle-consistency loss. The cycle-consistency loss and collective loss for this model are given in equations \ref{eq:4} and \ref{eq:5}, respectively.
\begin{multline}\label{eq:4}
\mathcal{L}_{cyc}(G,F) = \mathbb{E}_{x\sim p_{data(x)}}[\| F(G(x))- x\|_1] +\\ \mathbb{E}_{y\sim p_{data(y)}}[\| G(F(y))-y\|_1]
\end{multline}
\begin{multline}\label{eq:5}
\mathcal{L}(G,F, D_x, D_y) = \mathcal{L}_{GAN }(G, D_y, X, Y)+\\ \mathcal{L}_{GAN }(F, D_x, Y, X)+\mathcal{L}_{cyc}(G,F)
\end{multline}

\vspace{-0.5cm}

\subsection{Methodology}\label{methodology}
\begin{figure}[!h]
	\centering
	\includegraphics[clip,width=1\linewidth]{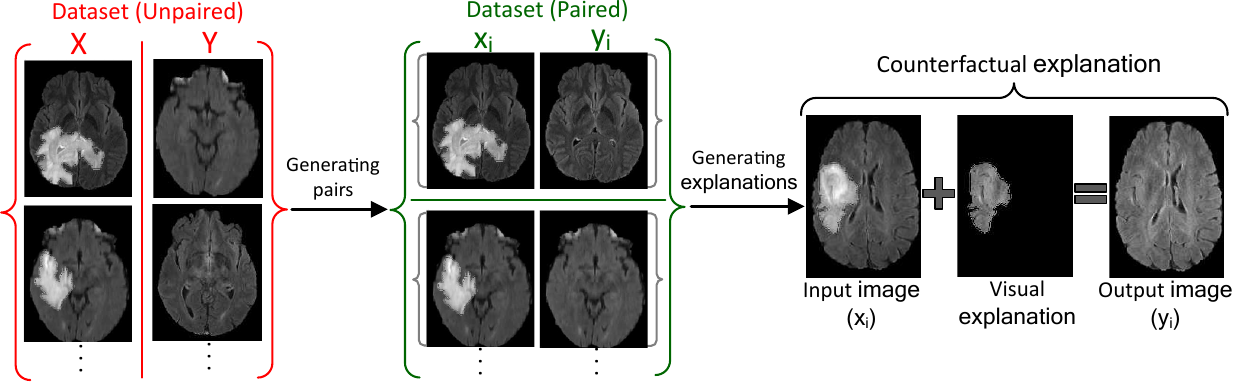}
	\captionsetup{justification=centering}
	\caption{Flow Diagram of CX-GAN}
	\label{Fig:4}
\end{figure}

The goal of this study is to devise a CX model for describing anomalous medical images with reference to their counterfactual normal images. In this work, this objective is divided into two tasks: 1) input-to-counterfactual translation (anomalous-to-normal translation), and 2) input-to-CX generation. The first task can be viewed as image-to-image translation from input domain to counterfactual domain. The second task can be viewed as a translation map that converts an input image to a counterfactual image. Although both tasks can be handled in a cascaded way, where in the first phase one can use image-to-image translation methods (such as CycleGAN) to produce counterfactual images of input images, and in second phase, one can use input counterfactual image pairs to produce a translation map for translating input into counterfactual images, as \cite{intro_36}. A systematic view of this approach is depicted in Figure \ref{Fig:4}.  However, two key disadvantages of this approach are: 1) two different networks are trained for different tasks; 2) performance of second task/network depends on the performance of first task/network. To overcome these issues, we propose an integrated model in the upcoming section.

\subsubsection{Integrated Model (CX-GAN)}
In this section, I present an integrated model for jointly learning to produce CX and CIs. It is assumed that a dataset of input X contains N images $\left\{X\right\}_{i=1}^N$ and counterfactual Y contains M images $\left\{Y\right\}_{i=1}^M$ is available; however, it is not in the form of pairs. The distribution of input images and counterfactual images is represented as $p_{data}(x)$ and $p_{data}(y)$, respectively. The model is built on CycleGAN. The counterfactual generation is accomplished using a generator $G:X\rightarrow$ y assisted by a discriminator $D_Y$. Optimization of G is achieved by an adversarial loss function $\mathcal{L}_G$ as defined below:
\begin{equation}
\begin{aligned}
\mathcal{L}_{GAN}(G, D_y, X, Y)
&= \mathbb{E}_{y\sim p_{data}(y)}[\log(D_y(y))] \\
&\quad + \mathbb{E}_{x\sim p_{data}(x)}
[\log(1-D_y(G(x)))] .
\end{aligned}
\end{equation}

Illustration of CI along with the input image is shown in Figure \ref{Fig:5}:

\begin{figure}[!h]
	\centering
	\includegraphics[clip,width=5cm]{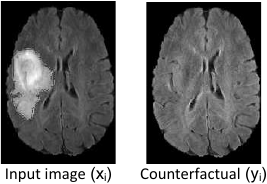}
	\captionsetup{justification=centering}
	\caption{Illustration of CIs}
	\label{Fig:5}
\end{figure}
In addition to translating input into a counterfactual image, I also want to produce a translation map (i.e., a change map) that, when added to an input image, produces a counterfactual image. CX is achieved using generator $G_M:Y+M(y)\rightarrow X$ with the assistance of discriminator $D_X$. Optimization of generator ``$G_M$" is done using the following loss function:
\begin{equation}
\begin{aligned}
\mathcal{L}_M(G_M, &D_x, Y, X)
= \mathbb{E}_{x\sim p_{data}(x)}
[\log(D_x(x))] \\
&\quad + \mathbb{E}_{y\sim p_{data}(y)}
[\log(1-D_x(G_M(y)+y))] .
\end{aligned}
\end{equation}

\begin{figure}[!h]
	\centering
	\includegraphics[clip,width=6.5cm]{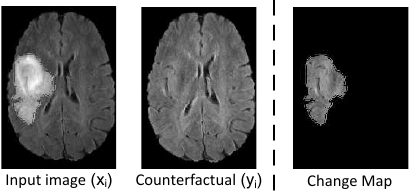}
	\captionsetup{justification=centering}
	\caption{Illustration of CX (Change Map)}
	\label{Fig:6}
\end{figure}

An illustration for CX (change map) by utilizing the generated pair (CI) and the input image is shown in Figure \ref{Fig:6}.

After CX we have to verify the translation between input and counterfactual domains. For this, we have to add a generated change map (counterfactual) to the normal images to ensure they look just like the input. To achieve this goal, we employed $\mathcal{L}_{cyclic}$ to ensure that when we add a map in the input image, it will look like the input image. $\mathcal{L}_{cyclic}$ is achieved using the following optimization function.
\begin{multline}
\mathcal{L}_{cyclic}(G,G_M)\ =\ \mathbb{E}_{x\sim p_{data(x)}}[\| G_M(G(x))-x\|_1] \\+ \mathbb{E}_{y\sim p_{data(y)}}[\|G(G_M(y)+y)-y\|_1]
\end{multline}
Objective functions for pair generation (CI generation), CX and cycle consistency loss are collectively optimized with the following mathematical function:
\begin{multline}
	\mathcal{L}_{CX-GAN}(G,G_M, D_x, D_y) = \mathcal{L}_{GAN}(G, D_y, X, Y)\\+\mathcal{L}_M(G_M, D_y, Y, X)+{\lambda\ \mathcal{L}}_{cyclic}(G,G_M)
\end{multline}

\begin{figure}[!h]
	\centering
	\includegraphics[clip,width=1\linewidth]{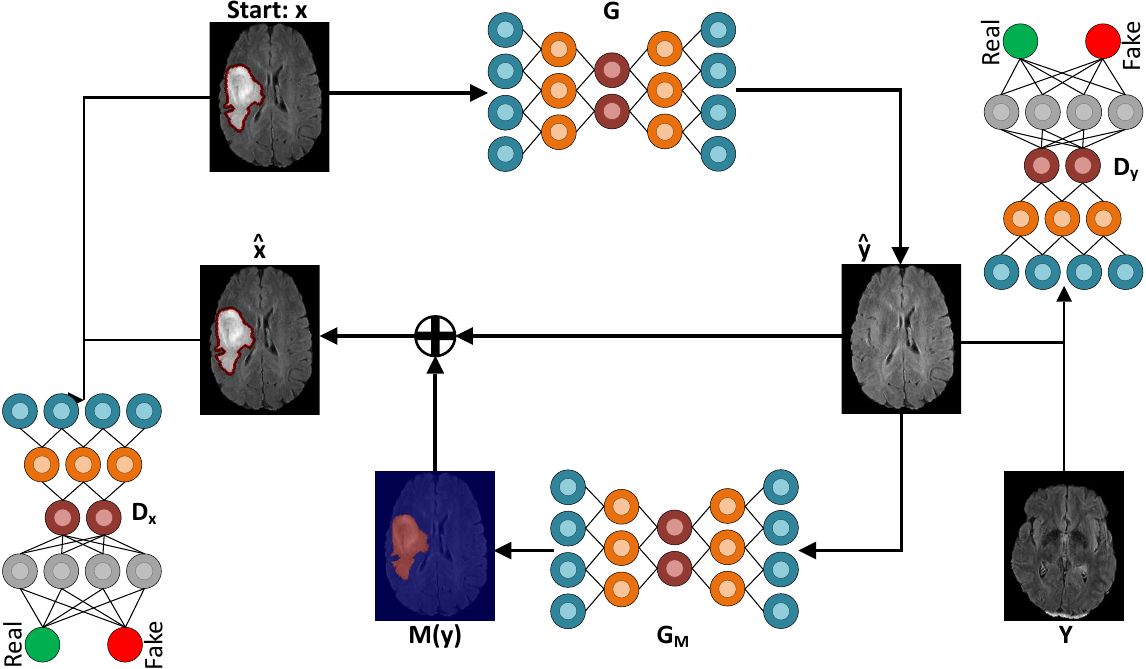}
	\captionsetup{justification=centering}
	\caption{System Diagram of CX-GAN}
	\label{Fig:7}
\end{figure}

The block diagram of the proposed model is depicted in Figure \ref{Fig:7}. The model consists of two generators, $G$ and $G_M$, and two discriminators, $D_y$ and $D_x$. The first generators G accepts the input image (tumorous image) and generate its respective normal pair (CI). The second generator, $G_M$, is used for the CX (change map). When the input to this generator is added to the generated counterfactual, the result will be the original image. Both generators learn to generate outputs with the assistance of their respective discriminators. 

\subsubsection{Network Architecture}
We have employed two types of networks in model; generator and discriminator. The generator consists of a shallow version of the standard U-Net, and the discriminator is a CNN. U-Net consists of an encoder and a decoder, with mirrored layers concatenated via shortcut connections. In contrast to other networks, U-Net employed skip connections that help preserve localization and contextual information during down- and up-sampling, respectively. Detailed specification of generator architecture (e.g. type, parameters) is shown in Table \ref{tab:generator_arch}. The generator M(x) is different from 1st generator utilized for CX by utilizing a change map, and this map can be obtained in two ways:
\begin{itemize}
	\item By appending the addition layer after the last convolution layer that utilizes input image and last layer before it. The change map is obtained from the last convolution layer, and the regenerated image is obtained from the output of the addition layer.
	\item In the second setting, we add the output of the generator and the input before applying the loss functions for optimization.
\end{itemize}
We employed both strategies and settings that yielded the same result because the additional layer did not employ any parameters that influence the learning of the generator.

\begin{table}[!h]
	\centering
	\caption{Generator's Architecture of CX-GAN}
	\label{tab:generator_arch}
	\resizebox{1\linewidth}{!}{%
		\begin{tabular}{|c|l|l|c|c|c|c|}
			\hline
			\textbf{\#} & \multicolumn{1}{c|}{\textbf{Layer}} & \multicolumn{1}{c|}{\textbf{Type}} & \textbf{\begin{tabular}[c]{@{}c@{}}Kernel Size \\ @ Filters \\ @ stride\end{tabular}} & \textbf{\begin{tabular}[c]{@{}c@{}}Output \\ Shape\end{tabular}} & \textbf{\begin{tabular}[c]{@{}c@{}}Param\\ \#\end{tabular}} & \textbf{\begin{tabular}[c]{@{}c@{}}Connected \\ to layer\#\end{tabular}} \\ \hline
			1 & Input & InputLayer & -- & 256, 256, 1 & 0 & -- \\ \hline
			2 & Convolution & Conv2D & 4@32@ 1 & 128, 128, 32 & 544 & 1 \\ \hline
			3 & Activation & LeakyReLU & -- & 128, 128, 32 & 0 & 2 \\ \hline
			4 & Normalization & Instance & -- & 128, 128, 32 & 2 & 3 \\ \hline
			5 & Convolution & Conv2D & 4@64@1 & 64, 64, 64 & 32832 & 4 \\ \hline
			6 & Activation & LeakyReLU & -- & 64, 64, 64 & 0 & 5 \\ \hline
			7 & Normalization & Instance & -- & 64, 64, 64 & 2 & 6 \\ \hline
			8 & Convolution & Conv2D & 4@128@1 & 32, 32, 128 & 131200 & 7 \\ \hline
			9 & Activation & LeakyReLU & -- & 32, 32, 128 & 0 & 8 \\ \hline
			10 & Normalization & Instance & -- & 32, 32, 128 & 2 & 9 \\ \hline
			11 & Convolution & Conv2D & 4@256@1 & 16, 16, 256 & 524544 & 10 \\ \hline
			12 & Activation & LeakyReLU & -- & 16, 16, 256 & 0 & 11 \\ \hline
			13 & Normalization & Instance & -- & 16, 16, 256 & 2 & 12 \\ \hline
			14 & Up-sampling & UpSampling2D & -- & 32, 32, 256 & 0 & 13 \\ \hline
			15 & Convolution & Conv2D & 4@128@1 & 32, 32, 128 & 524416 & 14 \\ \hline
			16 & Normalization & Instance & -- & 32, 32, 128 & 2 & 15 \\ \hline
			17 & Skip-connection & Concatenate & -- & 32, 32, 256 & 0 & 16 \& 10 \\ \hline
			18 & Up-sampling & UpSampling2D & -- & 64, 64, 256 & 0 & 17 \\ \hline
			19 & Convolution & Conv2D & 4@64@1 & 64, 64, 64 & 262208 & 18 \\ \hline
			20 & Normalization & Instance & -- & 64, 64, 64 & 2 & 19 \\ \hline
			21 & Skip-connection & Concatenate & -- & 64, 64, 128 & 0 & 20 \& 7 \\ \hline
			22 & Up-sampling & UpSampling2D & -- & 128, 128, 128 & 0 & 21 \\ \hline
			23 & Convolution & Conv2D & 4@32@1 & 128, 128, 32 & 65568 & 22 \\ \hline
			24 & Normalization & Instance & -- & 128, 128, 32 & 2 & 23 \\ \hline
			25 & Skip-connection & Concatenate & -- & 128, 128, 64 & 0 & 24 \& 4 \\ \hline
			26 & Up-sampling & UpSampling2D & -- & 256, 256, 64 & 0 & 25 \\ \hline
			27 & Convolution & Conv2D & 4@2@1 & 256, 256, 1 & 1025 & 26 \\ \hline
		\end{tabular}%
	}
\end{table}

Both discriminators in the model follow the same architecture. In contrast to the generator, the discriminator network contains a few layers because its task is to learn the decision boundary between real and fake inputs. Detailed specification for the discriminator network is given in Table \ref{tab:discriminator_arch}. 

\begin{table}[htbp]
	\centering
	\caption{Discriminator Architecture of CX-GAN}
	\label{tab:discriminator_arch}
 \resizebox{1\linewidth}{!}{%
	\begin{tabular}{|c|l|l|c|c|c|}
		\hline
		\textbf{\#} & \multicolumn{1}{c|}{\textbf{Layer}} & \multicolumn{1}{c|}{\textbf{Type}} & \textbf{\begin{tabular}[c]{@{}c@{}}Kernel size \\ @ Filters \\ @ Stride\end{tabular}} & \textbf{\begin{tabular}[c]{@{}c@{}}Output \\ Shape\end{tabular}} & \textbf{Param \#} \\ \hline
		1 & Input & InputLayer & -- & 256, 256, 1 & 0 \\ \hline
		2 & Convolution & Conv2D & 4*4 @ 64 @ 1 & 128, 128, 64 & 1088 \\ \hline
		3 & Activation & LeakyReLU & -- & 128, 128, 64 & 0 \\ \hline
		4 & Convolution & Conv2D & 4*4 @ 128 @ 1 & 64, 64, 128 & 131200 \\ \hline
		5 & Activation & LeakyReLU & -- & 64, 64, 128 & 0 \\ \hline
		6 & Normalization & Instance & -- & 64, 64, 128 & 2 \\ \hline
		7 & Convolution & Conv2D & 4*4 @ 256 @ 1 & 32, 32, 256 & 524544 \\ \hline
		8 & Activation & LeakyReLU & -- & 32, 32, 256 & 0 \\ \hline
		9 & Normalization & Instance & -- & 32, 32, 256 & 2 \\ \hline
		10 & Convolution & Conv2D & 4*4 @ 512 @ 1 & 16, 16, 512 & 2097664 \\ \hline
		11 & Activation & LeakyReLU & -- & 16, 16, 512 & 0 \\ \hline
		12 & Normalization & Instance & -- & 16, 16, 512 & 2 \\ \hline
		13 & Convolution & Conv2D & 4*4 @ 1 @ 1 & 16, 16, 1 & 8193 \\ \hline
	\end{tabular}
 }
\end{table}

\subsubsection{Network Training}
We follow the strategy of \cite{intro_36} to optimize the networks. The parameters of the generator and discriminator are updated alternatively. To restrain training oscillation, we update the discriminator’s parameters after having generated 5 images from the most recent generator, rather than a single image. All the networks are optimized with the ADAM optimizer. The learning rate and batch size are kept as 0.0002 and 1, respectively. The stopping criterion is chosen to be a patience of 10 epochs for validating the precision of generating a normal image pair. We train networks on a GPU-based desktop system with 128 GB RAM, Nvidia TitanX Pascal (12 GB VRAM) and a 10-core Intel Xeon processor.

\section{Experiments and Results}
\label{sec:4}

In this section, the results of the proposed method, along with the related methods, are presented. Both qualitative and quantitative results are produced to validate the performance of the proposed model. Details of evaluation techniques along with datasets are presented in section \ref{experimentalsetup}, and results are presented in section \ref{results}.

\subsection{Experimental Setup}\label{experimentalsetup}
The description of datasets and experimental settings used for training and evaluating the proposed model is discussed in this section.

\subsubsection{Datasets}
For testing purposes, three datasets have been used: (a) Synthetic dataset, (b) BraTS 2017, and (c) Shenzhen tuberculosis. Details of these dataset is given below:

\textbf{Synthetic dataset}

To validate the performance of the proposed method, a synthetic dataset is employed. It contains images of two categories; one is considered as normal, and the other is abnormal with class labels as ``0" and ``1" respectively. This dataset is generated by following the process discussed in \cite{intro_36}.

\begin{figure}[!h]
	\centering
	\includegraphics[clip,width=1\linewidth]{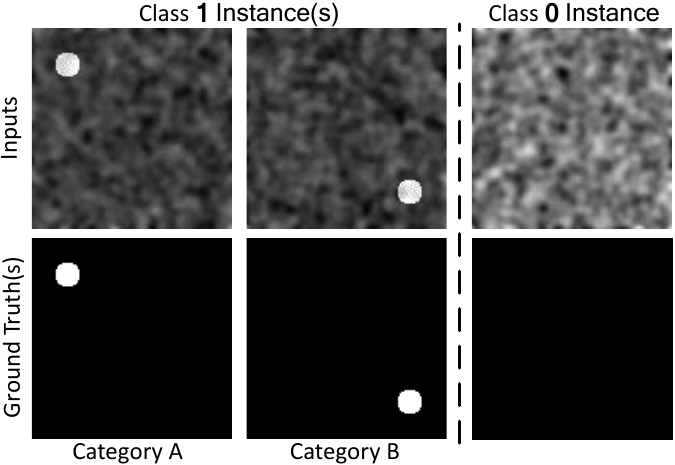}
	\captionsetup{justification=centering}
	\caption{Examples of synthetic data. Left of the dotted line are
		Samples of Class 1 (i.e. the disease class) and to the right of the dotted line are samples of Class 0 (i.e. the normal class). The upper row shows the input, and the bottom row shows the ground truth.}
	\label{Fig. 0}
\end{figure}

Images belonging to class ``0" contain noise that is drawn from a Gaussian distribution and is convolved with a noise filter. Class “1” images are generated the same way but with circles in them. Class ``1" instances are further divided into two categories; In category ``A", the circles are placed on the top left, and in category ``B", the circles are placed on the bottom right side of the image. Illustration of sample dataset instances is shown in Figure \ref{Fig. 0}.

\textbf{Brain Tumour dataset}

Brain tumour dataset is obtained from Multimodal Brain Tumour Segmentation (BraTS) 2017 challenge \cite{res_2, res_3}. This data contains both normal and abnormal images. MRI slices that contain the full brain are filtered from the dataset, which contains 463 normal and 3174 abnormal images. These filtered slices are then resized to $256 \times 256$. To boost the performance, run time augmentation is performed by resizing the image to $286 \times 286$ and then randomly cropping to $256 \times 256$ size.
\textbf{Tuberculosis dataset}

The Shenzhen Hospital tuberculosis chest X-rays (CXRs) dataset is also utilized for model evaluation, which contains both tuberculosis and non-tuberculosis images. This dataset is released by the National Institute of Health (NIH), US. This dataset contains 326 normal and 336 infected chest X-Rays \cite{res_4}. All images are resized to $256 \times 256$ and normalized using z-score normalization. For score calculation, ground truths for this dataset are obtained from the author of \cite{rw_6}.
\subsubsection{Evaluation}
The main objective of the proposed method is to learn CXs by measuring the change map between the generated CI and the input image. Qualitative and quantitative results for each dataset are discussed in the next section. For quantitative results, the peak value of these maps is obtained to generate a binary mask. DICE and intersection over union (IoU) are employed to calculate quantitative results for all datasets. Details of these matrices are given below:

\textbf{DICE: }Dice index is also used to find the similarity between two sets (or images), ranging between 0 and 1.

\begin{equation}
	DICE\ =\ \frac{2\times\ TP}{\left(FP+TP\right)+\left(FN+TP\right)}
\end{equation}

\textbf{Intersection over Union (IoU): }It tells how much of the desired object has been detected accurately. For example, if there are 0 wrong detected pixels (0 FN and 0 FP), the answer would be 1, indicating maximum match.
\begin{equation}
	IoU=\frac{TP}{TP+FN+FP}
\end{equation}

\subsection{Results}\label{results}
In this section, quantitative and qualitative results of the proposed model, along with related methods, are discussed.

\subsubsection{Experiments on synthetic dataset}
To validate the performance of the proposed model, a synthetic dataset is utilized. For quantitative results on this dataset, an additional loss is employed, named as normalized cross correlation (NCC). Quantitative results for this dataset are reported in Table \ref{tab:scores_syntheticdata}, depicting the relative supremacy of the proposed model over related models. The results of discriminative models are not reliable because they do not highlight the participating features at a fine-grained level.

\begin{table}[!h]
	\centering
	\caption{IoU, Dice Scores and NCC Scores of evaluated methods on synthetic data}
	\label{tab:scores_syntheticdata}
 \resizebox{1\linewidth}{!}{%
	\begin{tabular}{|l|c|c|c|c|}
		\hline
		\multicolumn{1}{|c|}{\multirow{2}{*}{\textbf{Method}}} & \multirow{2}{*}{\textbf{IoU Score}} & \multirow{2}{*}{\textbf{Dice Score}} & \multicolumn{2}{c|}{\textbf{NCC Score}} \\ \cline{4-5} 
		\multicolumn{1}{|c|}{} &  &  & \textbf{Mean} & \textbf{STD} \\ \hline
		CAM & 10.4 & 18.8 & 0.29 & 0.0025 \\ \hline
		gradCAM & 30.7 & 47 & 0.6 & 0.0482 \\ \hline
		VA-GAN & 87.2 & 92.8 & 0.92 & 0.0091 \\ \hline
		Integrated CX-GAN & 91.4 & 95.5 & 0.95 & 0.0061 \\ \hline
	\end{tabular}
 }
\end{table}

\begin{figure}[!h]
	\centering
	\includegraphics[clip,width=1\linewidth]{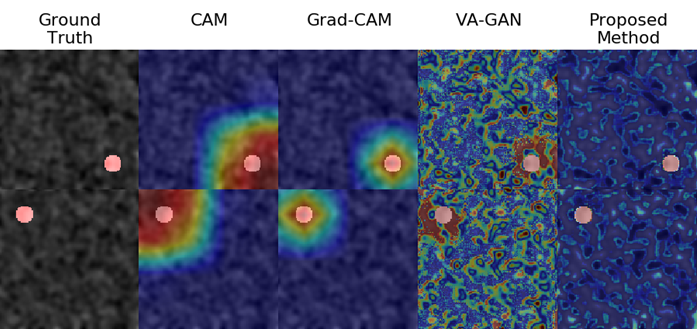}
	\captionsetup{justification=centering}
	\caption{Examples of visualization maps of compared methods on synthetic data.}
	\label{Fig. 1}
\end{figure}

Example output of our model and related methods is depicted in Figure \ref{Fig. 1}. Qualitative results show that discriminative model-based techniques (CAM and grad-CAM) are not able to provide visualizations at a fine-grained level. Qualitative results show that VA-GAN generates noise within the output map. This noise causes false positives, as this function is under a constraint function and allows mapping between unaligned images. These limitations affect the qualitative score of the VA-GAN method.

\subsubsection{Experiments on Shenzhen tuberculosis and BraTS dataset}
Quantitative results for both medical datasets are given in Table \ref{tab:score_brats_tb}. The scores of the proposed method are relatively high on both datasets. However, the model outperforms on the BraTS dataset because it contains anomalies that are easy to track. VA-GAN scored less due to the extra false positives because of translation between unaligned pairs. Discriminative methods perform very badly because they only focus on the minimal set of features in the input image.

\begin{table}[!h]
	\centering
	\caption{IoU and Dice Scores of evaluated methods on tuberculosis and BraTS datasets.}
	\label{tab:score_brats_tb}
 \resizebox{\linewidth}{!}{%
	\begin{tabular}{|l|c|c|c|c|}
		\hline
		\textbf{} & \multicolumn{2}{c|}{\textbf{Tuberculosis dataset}} & \multicolumn{2}{c|}{\textbf{BraTS dataset}} \\ \hline
		\multicolumn{1}{|c|}{Method} & \textbf{IoU Score} & \textbf{Dice Score} & \textbf{IoU Score} & \textbf{Dice Score} \\ \hline
		CAM & 19.67 & 28.56 & 30.8 & 45.1 \\ \hline
		gradCAM & 32.33 & 45.19 & 54.7 & 60.3 \\ \hline
		VA-GAN & 29.15 & 53.91 & 89.5 & 93.2 \\ \hline
		Integrated CX-GAN & 74.73 & 81.47 & 91.4 & 94 \\ \hline
	\end{tabular}
 }
\end{table}

\begin{figure}[!h]
	\centering
	\includegraphics[clip,width=0.8\linewidth]{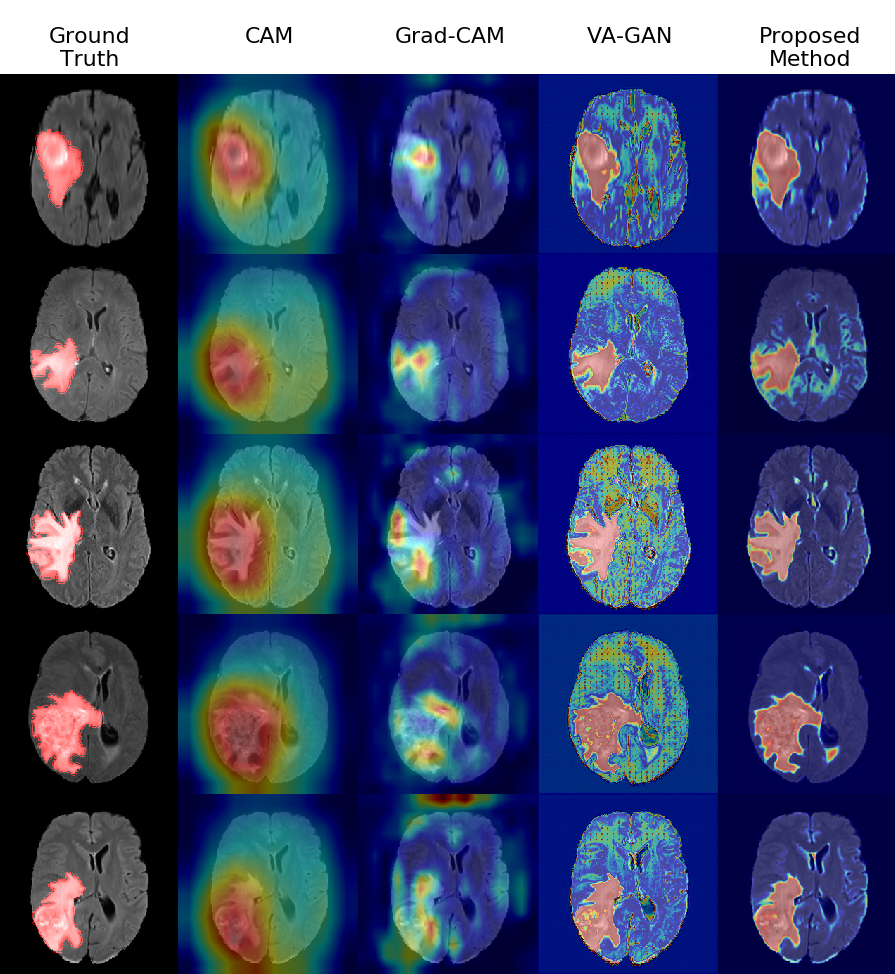}
	\captionsetup{justification=centering}
	\caption{Examples of visualization maps of the compared methods on BraTS data}
	\label{Fig. 2}
\end{figure}

Qualitative results of BraTS dataset are given in Figure \ref{Fig. 2}. These results show that our method produces results at a fine-grained level and the VA-GAN produces noise. CAM method covers the extra region other than the whole tumour, and grad-CAM covers a very small region compared to the actual tumorous region. Qualitative results show that the proposed method has not generated any noise along with the CX and captured the concerned region compared to related methods.

\begin{figure}[!h]
	\centering
	\includegraphics[clip,width=7cm]{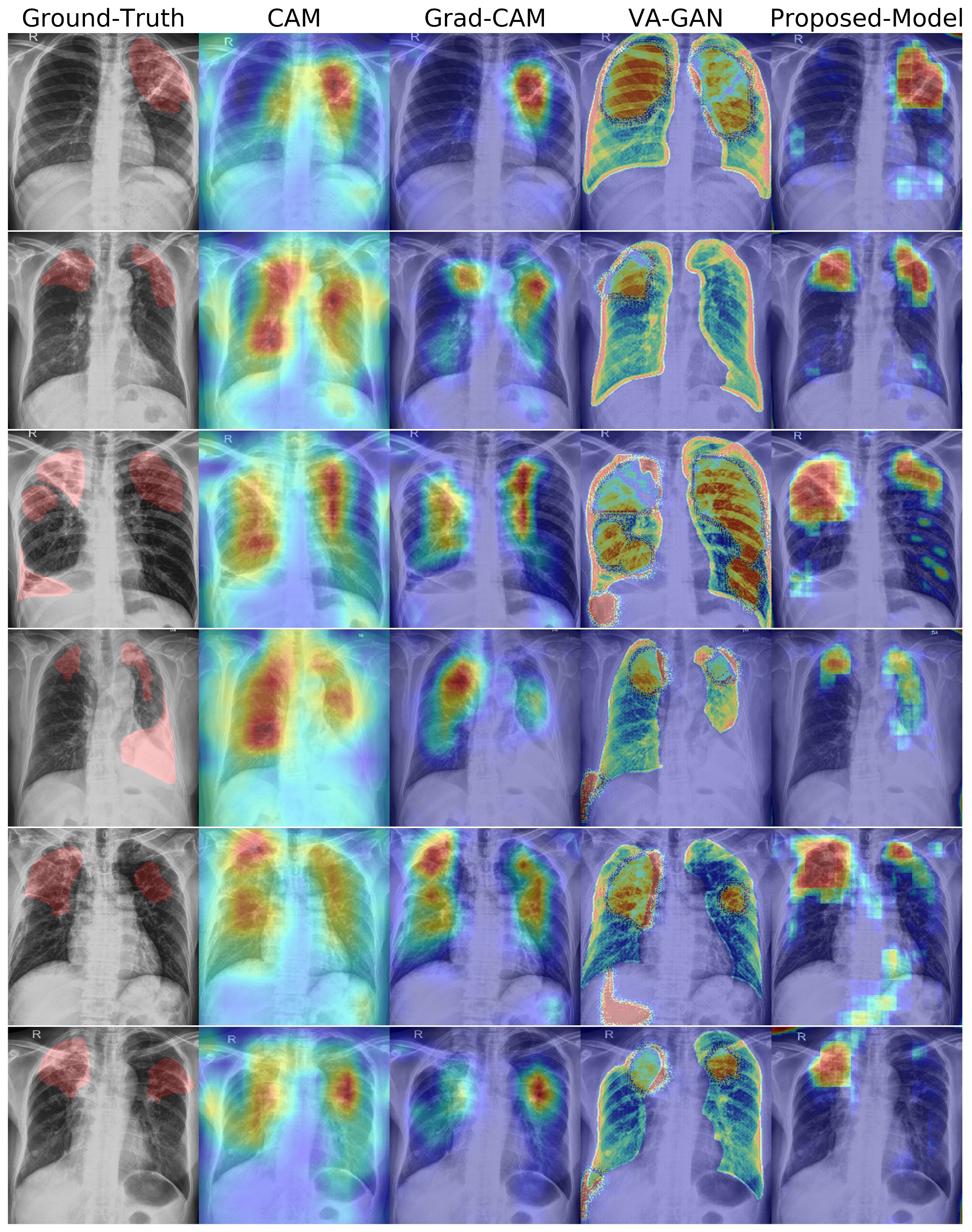}
	\captionsetup{justification=centering}
	\caption{Examples of visualization maps of compared methods on tuberculosis data.}
	\label{Fig. 3}
\end{figure}

Figure \ref{Fig. 3} shows the example output of our proposed method on Shenzhen dataset. CAM based methods do not cover the exact area of the concerned tumour and VA-GAN adds noise in the outpour image. Compared to related techniques our model generates plausible results.

\subsubsection{Discussion on generated CX}
\begin{figure}[!h]
	\centering
	\includegraphics[clip,width=1\linewidth]{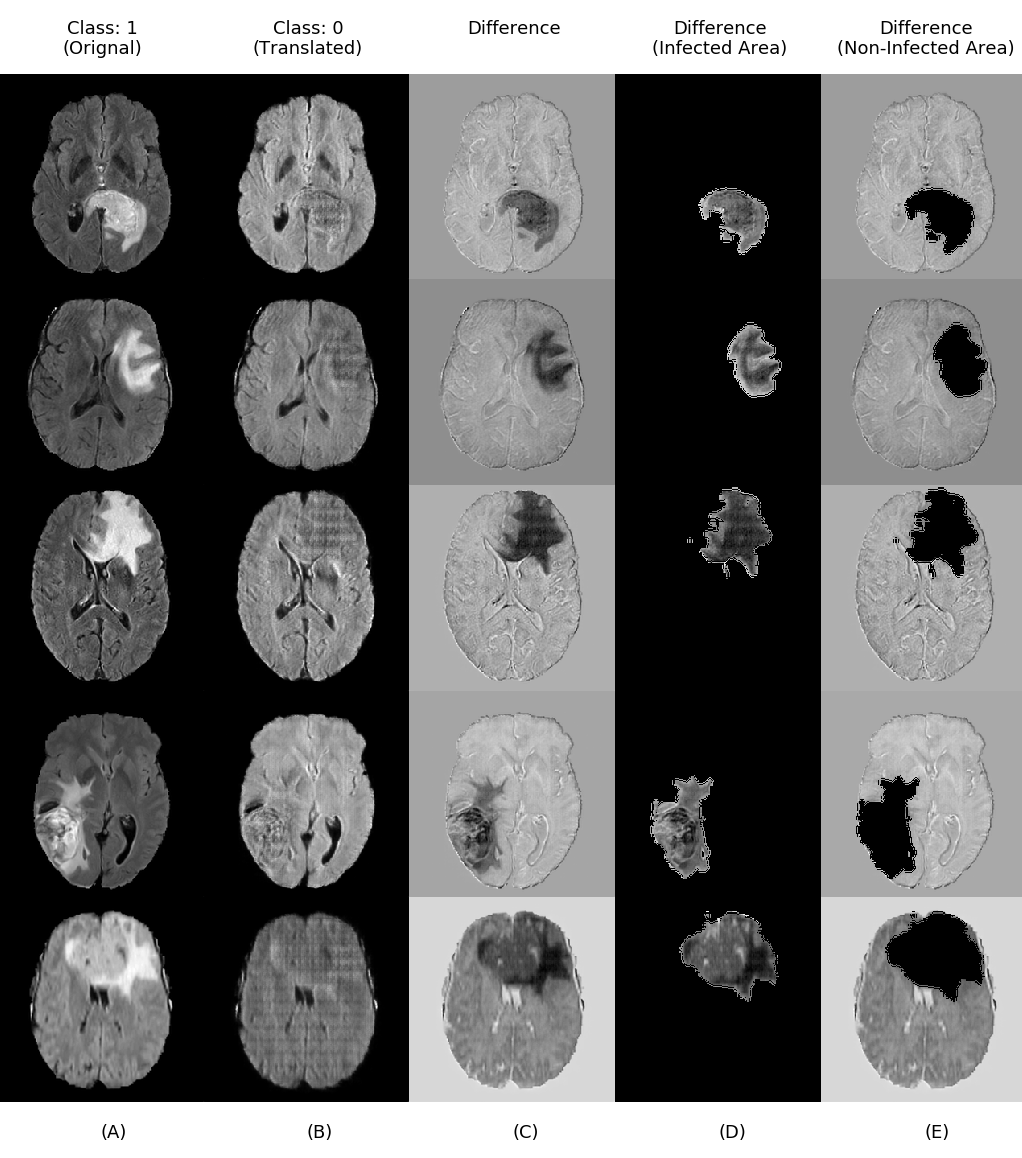}
	\caption{Illustration of the non-resemblance score measuring process. A) input image (i.e. tumorous image), B) generated CI (i.e. normal image), C) difference (or non-resemblance) image, D) separated tumorous region and E) separated normal region.}
	\label{Fig. 4}
\end{figure}

CXs and instances are generated using BraTS dataset. To evaluate the quality of generated counterfactual, a function calculates score of non-resemblance between input $(x_i)$ and generated CI $(y_i)$. The score is calculated separately for tumorous and normal regions. Non-resemblance is calculated using the difference between the tumorous and normal regions using the following,
\begin{multline}
	Tumerous\ regions\ =\ \ 1-\left(\frac{1}{N_i}\ \sum_{i}^{n_i}\left({y_i-\ x}_i\right)\ \right)\\ \ \ \ \therefore i:n_i\ where\ GT\ ==1
\end{multline}
\begin{multline}
	Normal\ regions\ =\ \ 1-\left(\frac{1}{N_j}\ \sum_{j}^{n_j}\left({y_j-\ x}_j\right)\ \right)\\ \ \ \ \therefore j:n_j\ where\ GT\ ==0
\end{multline}
Qualitative results for the generated instance with respect to the input image are shown in Figure \ref{Fig. 4}.

Table \ref{tab:Non_Resemblance_Score} shows the quantitative results for non-resemblance score between input and targeted domain. In contrast to VA-GAN, the proposed method scores high for translated tumorous regions and vice versa.

\begin{table}[!h]
	\centering
	\caption{Non-Resemblance Score on Brats data for generated pairs}
	\label{tab:Non_Resemblance_Score}
 \resizebox{\linewidth}{!}{%
	\begin{tabular}{|l|c|c|c|}
		\hline
		\multicolumn{1}{|c|}{\multirow{2}{*}{\textbf{Method}}} & \multicolumn{3}{c|}{\textbf{Non-Resemblance Score}} \\ \cline{2-4} 
		\multicolumn{1}{|c|}{} & \multicolumn{1}{l|}{Tumorous region} & \multicolumn{1}{l|}{Normal region} & \multicolumn{1}{l|}{Total Non-Resemblance} \\ \hline
		CX-GAN & 0.67 & 0.33 & 0.5 \\ \hline
		VA-GAN & 0.44 & 0.68 & 0.56 \\ \hline
	\end{tabular}
 }
\end{table}

\section{Conclusion}
\label{conclusion}
In this work, we developed a method for generating CXs and CI. These CX are produced using a change map between CI and its respective input. In contrast to existing methods, the proposed method is capable of producing plausible CI and fine-grained CXs. The model is built on a cycle-consistent GAN. The first generator is used for generating CIs, and the second one is utilized for the change map for producing CXs.

These CXs also help us to visualize the features learned by the model. These features are inferred from class-level labels and overcome the need for pixel-level labels. To validate the proposed model, three datasets are employed: the synthetic dataset, the tuberculosis dataset, and the brain tumour dataset.

\section{Future Work}
\label{futurework}
The following future directions can be used to extend our work.
\begin{itemize}
	\item This work focuses on generating the reasons for a particular decision. However, similar work can be done to make the model structure interpretable, whose decisions could be validated.
	\item A new CX layer could be designed that assists any black-box model to define/explain its own decision.
	\item Also, this model can be applied to other medical and safety-critical domains.
\end{itemize}

\bibliographystyle{abbrv}
\bibliography{main}

\end{document}